\documentclass[10pt,twocolumn,letterpaper]{article}

\usepackage{iccv}
\usepackage{times}
\usepackage{epsfig}
\usepackage{graphicx}
\usepackage{amsmath}
\usepackage{amssymb}
\usepackage{subcaption}
\usepackage{bbm}
\usepackage{algorithm}
\usepackage[noend]{algpseudocode}
\usepackage[toc,page]{appendix}
\usepackage{multirow}
\usepackage{booktabs}
\usepackage{CJKutf8}
\usepackage{makecell}

\usepackage{color, colortbl}
\definecolor{codegreen}{rgb}{0,0.6,0}
\definecolor{codegray}{rgb}{0.5,0.5,0.5}
\definecolor{codepurple}{rgb}{0.58,0,0.82}
\definecolor{backcolour}{rgb}{0.95,0.95,0.92}

\usepackage{listings}

\lstdefinestyle{mystyle}{
    commentstyle=\color{codegreen},
    keywordstyle=\color{magenta},
    numberstyle=\tiny\color{codegray},
    stringstyle=\color{codepurple},
    basicstyle=\ttfamily\footnotesize,
    breakatwhitespace=false,
    breaklines=true,
    captionpos=b,
    keepspaces=true,
    numbers=left,
    numbersep=5pt,
    showspaces=false,
    showstringspaces=false,
    showtabs=false,
    tabsize=2
}

\newcommand{\PreserveBackslash}[1]{\let\temp=\\#1\let\\=\temp}
\newcolumntype{C}[1]{>{\PreserveBackslash\centering}p{#1}}
\newcolumntype{R}[1]{>{\PreserveBackslash\raggedleft}p{#1}}
\newcolumntype{L}[1]{>{\PreserveBackslash\raggedright}p{#1}}

\usepackage[pagebackref=true,breaklinks=true,letterpaper=true,colorlinks,bookmarks=false]{hyperref}

\iccvfinalcopy 

\def\iccvPaperID{3} 

\begin{document}

\title{The Devil is in the Details: A Deep Dive into the Rabbit Hole of Data Filtering}

\author{Haichao Yu, Yu Tian, Sateesh Kumar, Linjie Yang, Heng Wang\\
ByteDance\\
{\tt\small \{haichaoyu,yutian.yt,sateesh.kumar1,linjie.yang,heng.wang\}@bytedance.com}
}

\maketitle

\begin{abstract}
The quality of pre-training data plays a critical role in the performance of foundation models. Popular foundation models often design their own recipe for data filtering, which makes it hard to analyze and compare different data filtering approaches. DataComp is a new benchmark dedicated to evaluating different methods for data filtering. This paper describes our learning and solution when participating in the DataComp challenge. Our filtering strategy includes three stages: single-modality filtering, cross-modality filtering, and data distribution alignment. We integrate existing methods and propose new solutions, such as computing CLIP score on horizontally flipped images to mitigate the interference of scene text, using vision and language models to retrieve training samples for target downstream tasks, rebalancing the data distribution to improve the efficiency of allocating the computational budget, \etc. We slice and dice our design choices, provide in-depth analysis, and discuss open questions. Our approach outperforms the best method from the DataComp paper by over 4\% on the average performance of 38 tasks and by over 2\% on ImageNet.

\end{abstract}

\section{Introduction} 
Multi-modal foundation models have been a major force in driving research progress on many topics in vision and language. Popular foundation models~\cite{radford2021learning,openai2023gpt4} are often built on large-scale image-text datasets~\cite{schuhmann2022laion,kakaobrain2022coyo-700m}. The quality of the image-text datasets~\cite{nguyen2022quality, santurkar2022caption} is critical for the performance of multi-modal models. The recipes for dataset curation have become the secret source behind the success of foundation models~\cite{radford2021learning,openai2023gpt4}. 
The DataComp challenge~\cite{datacomp} is created to systematically investigate and quantitatively evaluate different data filtering strategies. This paper describes our learning from participating in the DataComp challenge and discusses some of the open questions. Our goal is to provide more insights into dataset curation and inspire more research efforts in this critical direction.

\begin{figure}[t]
    \centering
    \includegraphics[width=0.8\linewidth]{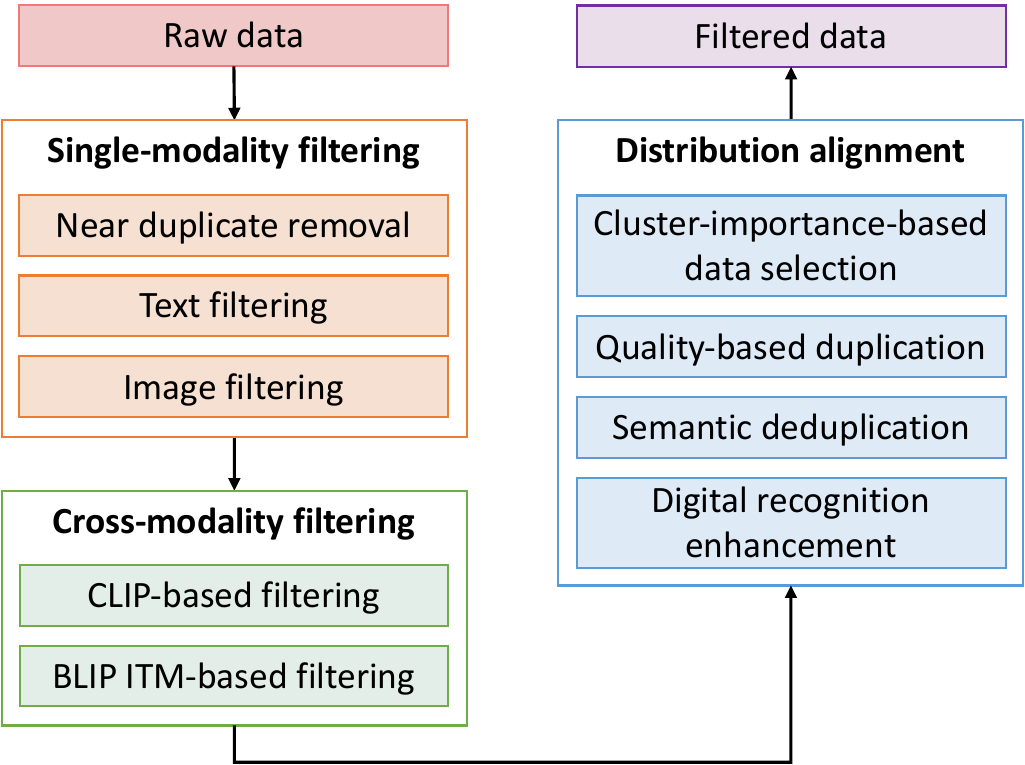}
    \caption{Our three-stage data filtering framework.}
    \label{fig:pipeline}
\end{figure}

Large-scale data curation is challenging and includes lots of technical details. Some early works~\cite{sharma2018conceptual,changpinyo2021conceptual, maini2023tmars} focus on the quality of individual samples. Other approaches~\cite{xie2023data,abbas2023semdedup} consider the distribution of the whole dataset and propose to improve the diversity of the dataset. Our data filtering framework is built upon previous approaches with new strategies proposed to improve the performance. Our framework consists of three major stages, as shown in Fig.~\ref{fig:pipeline}. \emph{Single-modality filtering} focuses on using signals from the image or text modality to filter out samples with obviously bad quality. \emph{Cross-modality filtering} mostly leverages CLIP-style models to select samples with good image-text similarity. The final stage, \ie, \emph{distribution alignment}, is to tailor the distribution of the pre-training dataset to fit the downstream tasks. As the filtering process 
continues stage by stage, we also tighten the criteria for selecting training samples.
Tab.~\ref{tab:curation_review} groups the major filtering techniques we experimented with for DataComp into four major categories, summarizes their effectiveness on DataComp, and compares our proposed methods with existing ones.

\begin{table}[t]
    \centering
    \small
    \begin{tabular}{c|p{7cm}}
    \toprule
        & Filtering methods \\
    \midrule
        \parbox[t]{2mm}{\multirow{15}{*}{\rotatebox[origin=c]{90}{Text}}}
            & [D] Safety detection (porngraphy, etc.) \cite{sharma2018conceptual,rae2021scaling,kakaobrain2022coyo-700m,changpinyo2021conceptual,jia2021scaling} \\
            & [U] PoS tag coverage/patterns \cite{sharma2018conceptual,changpinyo2021conceptual,li2021supervision,kakaobrain2022coyo-700m} \\
            & [E] Token repetition rate \cite{sharma2018conceptual,changpinyo2021conceptual,rae2021scaling} \\
            & [E] Ratios of: capitalized word; English word; symbols; words with $>1$ alphabetic characters \cite{sharma2018conceptual,li2021supervision,rae2021scaling} \\
            & [E] Removal of texts with rare tokens \cite{sharma2018conceptual,jia2021scaling} \\
            & [U] Predefined prefix/suffix patterns \cite{sharma2018conceptual,kakaobrain2022coyo-700m} \\
            & [E] Word or character length \cite{jia2021scaling,changpinyo2021conceptual,rae2021scaling,schuhmann2021laion,kakaobrain2022coyo-700m}\\
            & [U] Language detection \cite{alayrac2022flamingo,kakaobrain2022coyo-700m} \\
            & [E] Text importance estimation 
            \cite{brown2020language,xie2023data} \\
            & [U] Excluding alt-texts shared by $>10$ images \cite{jia2021scaling,kakaobrain2022coyo-700m} \\
            & [E] Documents should contain at least two words from \{\emph{the, be, to, of, and, that, have, with}\} \cite{rae2021scaling} \\
            & [U] Text deduplication \cite{schuhmann2021laion,rae2021scaling,brown2020language} \\
    \hline
        \parbox[t]{2mm}{\multirow{3}{*}{\rotatebox[origin=c]{90}{Image}}}
            & [U] Image deduplication \cite{alayrac2022flamingo,webster2023duplication} \\
            & [U] Image format, aspect ratio, color  \cite{sharma2018conceptual,changpinyo2021conceptual,jia2021scaling,schuhmann2021laion,kakaobrain2022coyo-700m,alayrac2022flamingo} \\
            & [E] Scene text recognition in image \cite{cao2023less,radenovic2023filtering} \\
    \hline
        \parbox[t]{2mm}{\multirow{4}{*}{\rotatebox[origin=c]{90}{Image-text}}}
            & [U] CLIP image-text similarity \cite{schuhmann2021laion,datacomp} \\
            & [E] Text-masked CLIP image-text similarity \cite{maini2023tmars,cao2023less,radenovic2023filtering} \\
            & [E] Larger CLIP variants (EVA-CLIP~\cite{EVA-CLIP}) \\
            & [O] Flipped-CLIP score\\
        \hline
        \parbox[t]{2mm}{\multirow{5}{*}{\rotatebox[origin=c]{90}{Dist. align.}}}
            & [U] Image-text joint deduplication \cite{schuhmann2021laion,kakaobrain2022coyo-700m} \\
            & [O] Cluster-importance-based data selection \\
            & [O] Quality-based duplication \\
            & [U] Semantic deduplication with image clustering \cite{abbas2023semdedup} \\
            & [O] Digit recognition enhancement \\
    \bottomrule
    \end{tabular}
    \caption{Overview of both existing and our proposed data filtering methods. 
    [D]: method performed by DataComp~\cite{datacomp}. [U]: existing method we used. [E]: existing method we evaluated, but does not work well. [O]: methods proposed by us for data filtering.}
    \label{tab:curation_review}
\end{table}

\section{Our approach}
\subsection{Single-modality Filtering}
The first stage uses either image or text information to remove samples with obviously bad quality. we loosely group them into the following three steps.

\noindent{\textbf{Near duplicate removal.}}
DataComp contains many near-identical image-text pairs, as shown in App.~\ref{app:removed_examples_near_identical}. 
We do both image and text deduplication to remove those near duplicates.
For image deduplication, we use an approximate KNN method from \cite{webster2023duplication}.  For text deduplication, samples are considered to be duplicates only if their texts are exactly the same. 
In this step, we only remove image-text pairs if their image and text are both duplicated. 
Please refer to App.~\ref{app:conser_dedup} for more details.

\noindent{\textbf{Text-based filtering.}}
As shown in Tab.~\ref{tab:curation_review}, we have experimented with many text-based filtering methods. We briefly describe several techniques that we found to be the most useful on DataComp.
{\it 1) Low-quality Part-of-Speech (PoS) patterns}. Inspired by \cite{sharma2018conceptual}, we generate PoS patterns for the text of each sample, and eliminate samples with PoS patterns indicative of low-quality texts, \eg, $\{\mathrm{PRON, ROOT, PUNCT}\}$. {\it 2) Low-quality high-frequency texts}. We collect all texts that occur more than $1,000$ times and manually remove the low-quality ones. {\it 3) Bad text patterns}. We qualitatively examine the text data from DataComp, and remove samples with specific patterns that are low-quality.
{\it 4) Non-English texts}. We remove non-English texts using language detectors. App.~\ref{app:text_cleaning} provides implementation details, text examples, and additional text-based rules we have tried and found unhelpful.

\noindent{\textbf{Image-based filtering.}}
We remove images with an aspect ratio $\frac{w}{h}$ outside the range of $[0.33, 3.33]$. Additionally, we discard images with a facial area ratio exceeding $0.4$, since human faces are blurred in DataComp.
We empirically found low-resolution images can be filtered by CLIP scores in the next stage, so we do not apply resolution-based rules in this step.

\subsection{Cross-modality Filtering}
\label{sec:cross_modality}
The second stage filters samples based on the similarity between the images and texts. CLIP score~\cite{radford2021learning} is the most widely used method for this purpose~\cite{schuhmann2022laion, datacomp, kakaobrain2022coyo-700m}.
The CLIP model has a very strong scene text recognition capability. If the image has scene texts overlapping with the caption, its CLIP score can be very high, even when the scene text does not reflect the semantic meaning of the image. Existing methods~\cite{cao2023less, radenovic2023filtering, maini2023tmars} often use text detection algorithms to mitigate this issue.

We propose to compute the CLIP score on horizontally flipped images (\ie, \textit{flipped-CLIP} score), instead of original images. This exploits the fact that CLIP models in \cite{radford2021learning} are trained without using horizontal flipping as data augmentation. Simply flipping the image can significantly reduce the scene text recognition capability from CLIP while preserving the semantic meaning of the image.
To compute \textit{flipped-CLIP} score, we employ the pre-trained ViT-L/14 model from \cite{radford2021learning}. App.~\ref{app:removed_examples_clip_flip} shows some examples with high CLIP scores but low flipped-CLIP scores. We find the flipped-CLIP score to be more selective for data filtering and has a higher peak performance than the CLIP score, as shown in App.~\ref{app:clip_vs_clip_flip}. 

In addition to the flipped-CLIP score, we also utilize the Image-Text Matching (ITM) score from BLIP~\cite{li2022blip}, which is trained with multiple objectives and different pre-training datasets.
We believe ITM-based filtering complements CLIP-based filtering, and uses ITM to remove a small portion of samples that flipped-CLIP score missed.
Please refer to App.~\ref{app:removed_examples_blip_itm} for visual examples.

\subsection{Distribution Alignment}
\label{sec:dist_align}
Unlike the previous two stages focusing on the quality of individual samples, the third stage aims to improve the overall data distribution, such as better dataset diversity (\emph{semantic deduplication}~\cite{abbas2023semdedup}), better efficiency for computational budget allocation (\emph{quality-based duplication}), better distribution alignment between the pre-training dataset and downstream tasks (\emph{cluster-importance-based data selection} and \emph{digit recognition enhancement}). We describe each step in the following paragraphs.

\noindent{\textbf{Cluster-importance-based data selection (CIDS).}}
The distribution of web-crawled data is often different from the desired distribution for model training. 
DSIR~\cite{xie2023data} aligns
the training data distribution with the target text corpora.
Note that DSIR is proposed for text datasets, where data distribution can be reliably estimated via word frequencies. We further extend DSIR to the image domain and use image clusters to estimate and align data distribution.
We leverage 100k image clusters provided by DataComp, measure the similarity of each cluster with downstream datasets, and assign each cluster a weight proportional to its similarity. We then resample the data so that samples get a higher representation in the filtered dataset if their clusters have a higher weight. Please refer to App.~\ref{app:data_select} for more details.

\noindent{\textbf{Quality-based duplication.}}
After all the previous steps, we manually check the quality of the remaining data, and find that there is still a significant difference among the quality of different samples. Given a fixed computational budget for training, we should allocate the budget to each sample based on their importance, instead of treating each sample equally. We propose to simply use the flipped-CLIP score to weight each sample, and repeat samples with higher weights multiple times in the filtered dataset so that they receive more computational budget during training (shown in Tab.~\ref{tab:results}).

More specifically, we sort samples $\{s_i\}|_{i=1}^{N}$ within each cluster from the previous step (\ie, CIDS) by their flipped-CLIP scores in ascending order.
We assign $w_1$ for the first sample $s_1$ and $w_2$ for the last sample $s_N$, and linearly interpolate the weights for the samples in between, \ie, assigning $\mathrm{round}((w_2 - w_1) \frac{i - 1}{N - 1} + w_1)$ for $s_i$.
The assigned weights $w$ represent the number of times we duplicate the sample.
We empirically found setting $w_1=1$ and $w_2=2$ yields the best performance.
Please refer to App.~\ref{app:quality_dup} for more details.

\begin{table*}[t]
    \centering
    \small
    \begin{tabular}{ll|llcccccc}
    \toprule
        \multicolumn{2}{c|}{\multirow{2}{*}{Methods}} & \multirow{2}{*}{Dataset size} & \multicolumn{2}{c}{38 tasks} & IN & IN shifts & VTAB & Retr. & Digit \\
        \cline{4-10}
        & & & avg. & best & avg. & avg. & avg. & avg. & avg.\\
        \midrule
        \multicolumn{2}{l|}{No filtering~\cite{datacomp}} & 128M & 0.258 & - & 0.176 & 0.152 & 0.259 & 0.219 & - \\
        \hline
        \multicolumn{2}{l|}{Image-based $\cap$ CLIP score (L/14 30\%)~\cite{datacomp}} & 14M & 0.328 & - & 0.297 & 0.239 & 0.346 & 0.231 & - \\
        \midrule
        \multicolumn{2}{l|}{Single-modal filtering} & 96M & 0.274 {\scriptsize ($\pm$0.0026)} & 0.280 & 0.198 & 0.170 & 0.271 & 0.234 & 0.089 \\
        \hline
        \multicolumn{2}{l|}{+Cross-modal filtering} & 54M & 0.322 {\scriptsize ($\pm$0.0037)} & 0.328 & 0.265 & 0.223 & 0.320 & \textbf{0.270} & 0.106 \\
        \hline
        \multirow{4}{*}{+Dist. align.} & +CIDS & 24M & 0.352 {\scriptsize ($\pm$0.0027)} & 0.357 & 0.318 & 0.258 & 0.363 & 0.260  & 0.141 \\ 
        & +Quality-based dup. & 24M {\scriptsize (36M)} & 0.353 {\scriptsize ($\pm$0.0034)} & 0.358 & 0.323 & 0.261 & 0.365 & 0.260  & 0.122 \\ 
        & +Semantic dedup. & 22M {\scriptsize (33M)} & 0.359 {\scriptsize ($\pm$0.0031)} & 0.365 & \textbf{0.329} & \textbf{0.266} & 0.373 & 0.263 & 0.122 \\
        & +Digit recog. enhancement & 23M {\scriptsize (36M}) & \textbf{0.362} {\scriptsize ($\pm$0.0040)} & \textbf{0.371} & 0.318 & 0.259 & \textbf{0.377} & 0.258 & \textbf{0.304} \\
    \bottomrule
    \end{tabular}
    \caption{Results of applying each filtering method from Fig.~\ref{fig:pipeline} on the medium track of DataComp. 
    \textit{Dataset size} is the number of unique data samples.  
    The sizes in parentheses are the number of samples after duplication. Results are averaged over $10$ runs. For the 38-task average performance, we provide their standard deviations in parentheses. Best results are in \textbf{bold}.} 
    \label{tab:results}
     \vspace{-1em}
\end{table*}

\noindent{\textbf{Semantic deduplication.}}
By analyzing the clustering results from CIDS, 
we find the number of samples per cluster is very unbalanced
(as shown in App.~\ref{app:semantic_dedup_cluster_imbalance}). This observation motivates us to diversify the filtered dataset by removing  
semantically similar images~\cite{abbas2023semdedup}. Using the unique data samples from the previous step (\ie, \emph{quality-based duplication}), we do k-means clustering to obtain $k$ image clusters with CLIP ViT-L/14 embeddings. We set $k=100\mathrm{k}$ in our experiments. Then, within each cluster, we calculate 
the cosine similarities $S$ among all the image pairs.
We divide the images within the cluster into groups using a threshold on the similarity, \ie, image $I_i$ and $I_j$ are in the same group if $S_{i,j} > thre_{sem}$.
Unlike~\cite{abbas2023semdedup} keeping the image farthest from its cluster centroid, we select the sample with the highest CLIP score and remove the others.
In App.~\ref{app:semdedup}, we show additional details and visual examples. Note that \emph{quality-based duplication} and \emph{semantic deduplication} are independent, and changing their order does not impact the performance. 

We also experimented with using image-text joint embedding for semantic deduplication in order to better cluster the image-text samples.
However, it does not produce better results compared to using image embedding alone. For further details, please refer to App.~\ref{app:semdedup}.

\noindent{\textbf{Digit recognition enhancement.}}
The 38 downstream tasks proposed by DataComp~\cite{datacomp} include MNIST~\cite{lecun1998mnist} and SVHN~\cite{netzer2011reading} datasets for digit recognition. We observe that the baselines proposed in \cite{datacomp} perform poorly on the digit recognition tasks. We hypothesize that a significant portion of data with digits in them is filtered out when using CLIP score
for filtering. To alleviate this issue, we identify images
with digits in them from the whole dataset and add these samples to the filtered dataset.

Specifically, we create text prompts to identify images with 
digits and use the BLIP-2~\cite{li2022blip} model to scan the whole dataset with the designed prompts. Based on the responses from the BLIP-2 model, we can identify samples
that are classified as useful for digit recognition. We find the BLIP-2 model is very effective in finding images containing digits. Please refer to App.~\ref{app:added_examples_blip2} for more details on the method and qualitative examples. As we show later, obtaining downstream task-aligned data in this manner leads to an $18\%$ improvement in digit recognition performance in our experiments.

\section{Experimental Results}
\label{sec:results}
We experimented with our proposed filtering methods
on the medium filtering track of the DataComp challenge. Out of 128M samples, we successfully downloaded 120.7M images. 

\noindent{\textbf{Implementation details.}} We follow the training protocols provided by DataComp. All our models are trained with 32 Nvidia A100 GPUs. 
For \emph{cross-modality filtering}, we remove data with a flipped-CLIP score lower than $0.19$. For CIDS, we set the threshold $thre_{imp}$ to $0.72$. 
We use the training sets of the $22$ downstream tasks whose train-test splits are available for estimating importance weights.
For \emph{semantic deduplication}, we set $thre_{sem}=0.04$.

We evaluate our models following the same setting from DataComp, as shown in Tab.~\ref{tab:results}. We also report the average accuracy on two
datasets (MNIST~\cite{lecun1998mnist} and SVHN~\cite{netzer2011reading}) to showcase the improvement from digit recognition enhancement.
We find there is a noticeable variance in results from the 38
tasks when using the same training data\footnote{The variance persists despite using a fixed training seed: \url{https://github.com/mlfoundations/datacomp/blob/main/slurm_train.sh\#L33}}.
For a more reliable measure of performance,  
we train at least ten models for each filtering setting and report their average performance. We also report the best performance of the ten models in Tab.~\ref{tab:results}.

\noindent{\textbf{Result analyses.}} Tab.~\ref{tab:results} compares our method with the baseline (\ie, no filtering) and best results from \cite{datacomp}. First, each step of our framework progressively improves the performance on the 38 tasks.
\emph{Single-modal filtering} improves the average performance from $0.258$ to $0.274$ by removing $20.5\%$ of data (\ie, from 120.7M to 96M), whereas \emph{cross-modal filtering} further boosts the performance to $0.322$ by cutting down the dataset size to 54M. With \emph{distribution alignment}, we achieve the best performance of $0.362$ and the dataset size is reduced by another half (\ie, from 54M to 23M).
Since the computational budget is the same for all the methods, this clearly demonstrates the effectiveness of our proposed framework. 
Each step in the framework can bring extra value and is complementary to each other.

Second, aligning the pre-training dataset distribution with target downstream tasks is really important. We further break down the improvements from each step of \emph{distribution alignment} in Tab.~\ref{tab:results}. The ImageNet accuracy is improved by $5.3\%$ (\ie, from $0.265$ to $0.318$), and digit recognition by $19.8\%$ (\ie, from $0.106$ to $0.304$). The improvement on digit recognition is achieved by only adding $1.3\%$ of total training data (\ie, 1.6M samples). Given the fixed computational budget, identifying training samples that are directly related to the target task is the most cost-effective.

Third, we also observe inconsistency among different tasks. The performance of different tasks can be competing with each other, since the computational budget is fixed. For example, adding 1.6M digit training data improves digit recognition by $18.2\%$, but the accuracy on ImageNet is decreased by $0.9\%$. We also notice that some of the 38 tasks are particularly challenging, and the accuracy on them is almost random guess, including CLEVR Counts and Distance~\cite{johnson2017clevr}, and Rendered SST2~\cite{zhai2019vtab}. They introduce lots of variance to the average performance of the 38 tasks. App.~\ref{app:ours_vs_datacomp} shows some samples selected by our approach but not included by the best method from \cite{datacomp}.

\section{Discussion}

Besides the methods we described in the previous sections, we also experimented with other ideas that did not work well during the DataComp challenge. There are also many open questions that we do not have answers yet. We briefly discuss some of them that may be worth further investigation in this section.

\noindent{\textbf{Images with scene text.}}
Several previous works~\cite{radenovic2023filtering,cao2023less,maini2023tmars} demonstrate that images with lots of scene texts can be low-quality, especially if the scene texts overlap with the caption of the image. We propose the flipped-CLIP score to mitigate this issue.
However, scene text recognition can be critical for many downstream tasks~\cite{openai2023gpt4}.
Samples with scene text can be very useful in order to improve the performance on tasks that require scene text recognition~\cite{liu2023hidden}. We should strike a better balance when filtering images with scene texts, instead of simply removing them.

\noindent{\textbf{Does a stronger model lead to a better filtering method?}}
The CLIP score is widely used for data filtering~\cite{schuhmann2021laion,datacomp}. One may hypothesize the filtering method will be more effective if we use a better model than CLIP. We tested the EVA02-CLIP-bigE-14-plus model~\cite{EVA-CLIP} to verify this hypothesis. 
Despite its stronger performance than CLIP ViT-L/14, it does not lead to a better filtering method
(Refer to App.~\ref{app:clipb_clipl_evaclipe} for details.). Similar results are presented in \cite{datacomp} (Tab.~24-27). It requires more investigation to better understand the relationship between the performance of the model and the corresponding data filter. 

\noindent{\textbf{Generalization of data filtering methods.}}
Data filtering methods may not generalize well to different data distributions. 
We observed inconsistent results when applying solutions developed on the medium track to the large track of DataComp.
Similar trends also exist when comparing other methods on different tracks\footnote{\url{https://www.datacomp.ai/leaderboard.html}}.
These findings underscore the challenge of developing data filtering solutions that can cope with different data distributions and scales.

{\small
\bibliographystyle{ieee_fullname}
\bibliography{src_references}
}

\clearpage

\appendix

\section{Appendix}
\subsection{Near-identical Visual Examples}
\label{app:removed_examples_near_identical}
\begin{figure}[!ht]
  \begin{subfigure}{0.32\linewidth}
  \includegraphics[width=\linewidth, height=\linewidth]{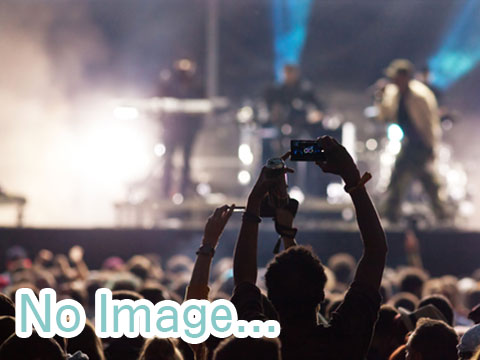}
  \caption{``株式会社フルキャスト　中部支社　静岡営業課/FN0926H-ACのアルバイト情報''. C-f: 0.129. N: 25.}
  \end{subfigure}
 \begin{subfigure}{0.32\linewidth}
  \includegraphics[width=\linewidth, height=\linewidth]{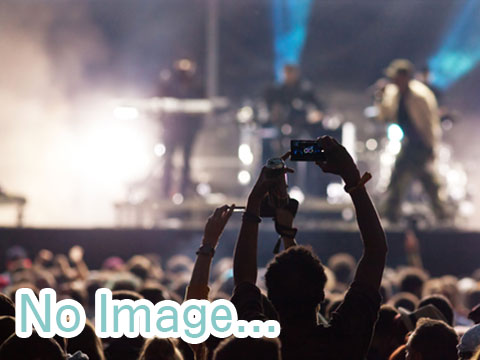}
  \caption{``株式会社フルキャスト　中部支社　静岡営業課/FN0926H-ACのアルバイト情報''. C-f: 0.129.}
  \end{subfigure}
   \begin{subfigure}{0.32\linewidth}
  \includegraphics[width=\linewidth, height=\linewidth]{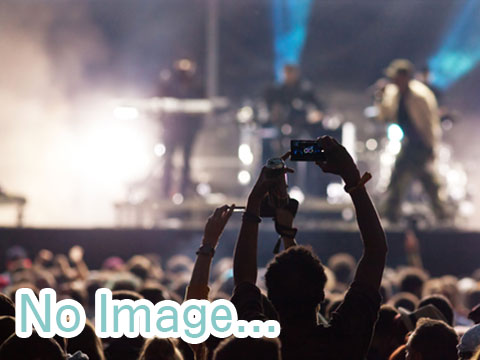}
  \caption{``(株)フルキャスト 埼玉支社 川口センター/FN1005F-9Fのアルバイト情報''. C-f: 0.112.}
  \vspace{1.1em}
  \end{subfigure}

  \begin{subfigure}{0.32\linewidth}
  \includegraphics[width=\linewidth, height=\linewidth]{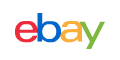}
  \caption{``Franklin Brass Fbdprh5-Mn-C Double Prong Robe Hook 5-Pack Matte Nickel 5 Piece''. C-f: 0.149. N: 970.}
  \end{subfigure}
 \begin{subfigure}{0.32\linewidth}
  \includegraphics[width=\linewidth, height=\linewidth]{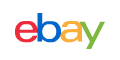}
  \caption{``Chain Saw Service Manual: 10th Edition Review and Comparison''. C-f: 0.149.}
  \vspace{1.11em}
  \end{subfigure}
  \begin{subfigure}{0.32\linewidth}
  \includegraphics[width=\linewidth, height=\linewidth]{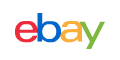}
  \caption{``Spenco Total Support Max Shoe Insoles, Women's 7-8/Men's Review and Comparison''. C-f: 0.153.}
  \end{subfigure}

  \begin{subfigure}{0.32\linewidth}
  \includegraphics[width=\linewidth, height=\linewidth]{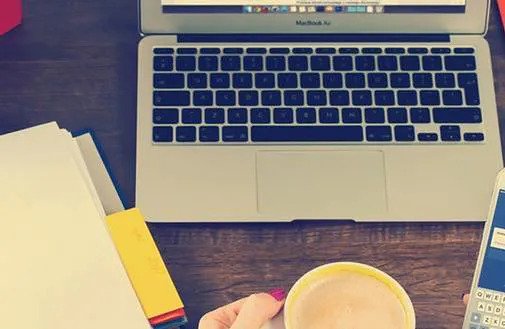}
  \caption{``Synerga.fund podpisala term sheet w sprawie polqczenia z All In! Games''. C-f: 0.145. N: 5.}
  \end{subfigure}
  \begin{subfigure}{0.32\linewidth}
  \includegraphics[width=\linewidth, height=\linewidth]{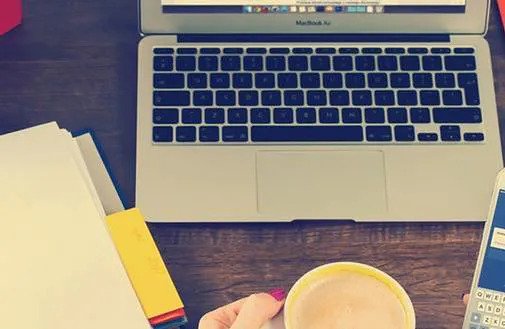}
  \caption{``Synerga.fund podpisala term sheet w sprawie polqczenia z All In! Games''. C-f: 0.145.}
  \vspace{1.11em}
  \end{subfigure}
  \begin{subfigure}{0.32\linewidth}
  \includegraphics[width=\linewidth, height=\linewidth]{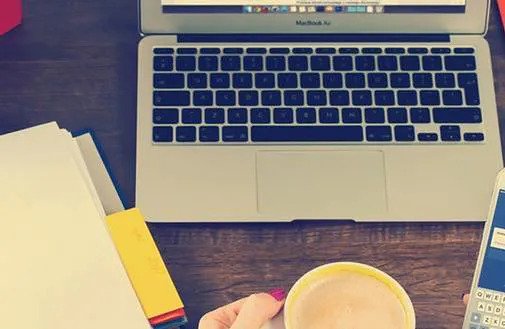}
  \caption{``Synerga.fund podpisala term sheet w sprawie polqczenia z All In! Games''. C-f: 0.145.}
  \vspace{1.11em}
  \end{subfigure}
  
  \caption{Near-identical duplicate samples. Samples in a row come from the same group. C-f: flipped-CLIP score. N: number of samples within the group.}
  \label{fig:near_identical}
\end{figure}

In Fig.~\ref{fig:near_identical}, we show visual examples detected by the \emph{near duplicate removal} step.

\begin{CJK}{UTF8}{ipxm}

\end{CJK}

\subsection{More Details on Near Duplicate Removal}
\label{app:conser_dedup}
\begin{algorithm}[t]
\caption{Near duplicate removal.}\label{alg:dedup}
\hspace*{\algorithmicindent} \textbf{Input}: raw dataset $\mathcal{D}$, inverted index $I$. \\
\hspace*{\algorithmicindent} \textbf{Output}: filtered dataset $S$.

\begin{algorithmic}[1]

\State $S \gets \emptyset$
\State $S_{remove} \gets \emptyset$

\For{$x \in \mathcal{D}$}
\If{$x \notin S_{remove}$}
    \State $y_{n} \gets \mathrm{getImageDuplidates}(I, \tau, x)$ 
    \If{$y_{n} = \emptyset$}
        \State $S \gets S \cup \{x\}$
    \Else
        \For{$y \in y_n$}
            \If{$\mathrm{getCLIPScore}(y) \leq \gamma$}
                \State $S_{remove} \gets S_{remove} \cup \{y\}$
            \EndIf
        \EndFor
        \If{$\mathrm{getCLIPScore}(x) > \gamma$}
            \State $S \gets S \cup \{x\}$
            \For{$y \in y_n$}
                \If{$\mathrm{getCLIPScore}(y) > \gamma$ \newline
                    \hspace*{8em} \textbf{and} $t_x = t_y$} \Comment{Text dedup.}
                    \State $S_{remove} \gets S_{remove} \cup \{y\}$
                \EndIf
            \EndFor
        \EndIf
    \EndIf
\EndIf
\EndFor
\Return $S$
\end{algorithmic}
\end{algorithm}

\noindent{\textbf{Image deduplication.}} Following \cite{webster2023duplication}, we build inverted files with Product Quantization on the compressed CLIP embeddings to index all the images. Given a query image $x$, we find its approximate $k$ nearest neighbors $\{y_{i}\}|_{i=1}^{k}$ by searching through the inverted files. Subsequently, a neighbor $y$ is flagged as a duplicate of $x$ if it satisfies the condition:
$\frac{d_{ADC}(x, x) - d_{ADC}(x, y)}{d_{ADC}(x, x)} < \tau$, where $d_{ADC}(x, y)=||x - \mathrm{quantizer}(y) ||$ represents the distance between $x$ and the quantized version of $y$. We use the IndexIVFPQ implementation available in FAISS~\cite{johnson2019billion} to perform this algorithm. Here are the detailed configurations:
\begin{itemize}
    \setlength\itemsep{0em}
    \item Number of clusters for k-means: $nlist=65,536$.
    \item Number of chunks for Product Quantization: $M=4$.
    \item Number of bits for each index in Product Quantization: $nbits\_per\_idx=8$.
    \item Number of clusters to probe: $nprobe=2$.
    \item Number of returned nearest neighbors: $k=1,024$.
    \item The duplication threshold: $\tau=0.03$.
    \item CLIP score threshold in Alg.~\ref{alg:dedup}: $\gamma=0.19$.
\end{itemize}

\noindent{\textbf{Near duplicate removal.}}
For the whole pipeline of near duplicate removal, refer to  Alg.~\ref{alg:dedup}.

\subsection{More Details on Text-based Filtering}
\label{app:text_cleaning}
\begin{figure}[t]
  \includegraphics[width=\linewidth]{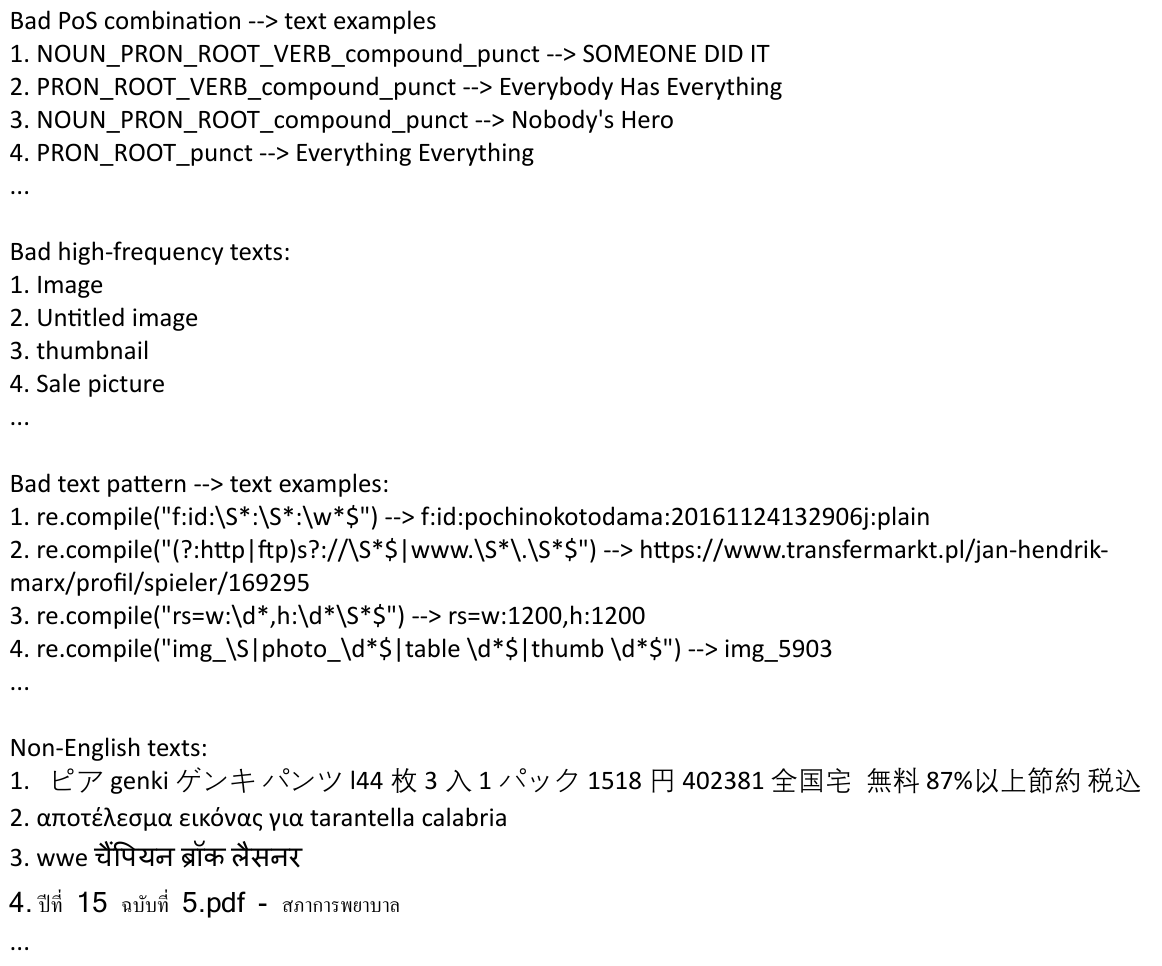}
  \caption{Text examples filtered by each text filtering rule.}
  \label{fig:removed_text_examples}
\end{figure}

In this section, we first provide more implementation details for our text-based filtering rules as follows:
\begin{itemize}
    \setlength\itemsep{0em}
    \item {\it Part-of-Speech (PoS) combination}. Prior work~\cite{sharma2018conceptual} shows that a well-formed caption should have a high POS tag coverage. In practice, we find this will remove a lot of useful data as well. Instead, we consider \emph{non-repeatitive POS combinations}. Specifically, we get PoS tags for each DataComp caption using spaCy~\cite{spacy2}. Then the non-repeatitive tags are sorted and concatenated as a string-form PoS pattern. We visualize examples from all possible PoS patterns, and remove texts corresponding to low-quality patterns.
    \item {\it High-frequency texts}. We collect all texts that appear more than $1,000$ times and remove meaningless texts.
    \item {\it Bad text patterns}. We run fasttext language detector~\cite{joulin2016bag} on all texts, then qualitatively examine examples with scores less than $0.2$ for all languages. We find some patterns are shared by many low-quality texts, and designed rules to detect and remove them.
    \item {\it Non-English data}. We remove non-English texts with a fasttext score above a threshold. We set different thresholds for different non-English languages.
\end{itemize}
In Fig.~\ref{fig:removed_text_examples}, we present some caption examples deleted by these rules.

In addition to these rules, we also conduct experiments with other text filtering criteria, including
\begin{itemize}
    \setlength\itemsep{0em}
    \item { \it Word count}. A good caption should contain a sufficient number of words.
    \item { \it Character count}. A good caption should contain a sufficient number of characters.
    \item { \it Average word length}. A good caption's average word length should be larger than a specific threshold.
    \item { \it Word alphabetic ratio}. A good caption should meets the criterion $\frac{num\_alpha\_tokens}{num\_tokens}>\mathrm{threshold}$. alpha\_token: a token that contains at least one alphabetic character.
    \item { \it Unique token ratio}. A good caption should meets the criterion $\frac{num\_unique\_tokens}{num\_tokens} > \mathrm{threshold}$.
    \item { \it Synset-based rules}. We collect a synset list based on class names of the downstream tasks, and filter out the texts which have no overlap with the list.
    \item { \it Text importance estimation}. Following~\cite{xie2023data}, we estimate importances on DataComp texts, using as targets the downstream class names and the captions from Conceptual-Captions~\cite{sharma2018conceptual} .
\end{itemize}
However, our ablation studies reveal that these additional rules contribute only marginal or negligible improvements. Consequently, we have opted not to include it in our final pipeline.

\subsection{Visual Examples with Low flipped-CLIP Scores}
\label{app:removed_examples_clip_flip}
\begin{figure}[t]
  \begin{subfigure}{0.32\linewidth}
  \includegraphics[width=\linewidth, height=\linewidth]{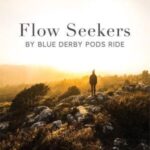}
  \caption{``Flow Seekers By Blue Derby Pods Ride''. C-f: 0.144. C: 0.478.}
  \vspace{2.2em}
  \end{subfigure}
  \begin{subfigure}{0.32\linewidth}
  \includegraphics[width=\linewidth, height=\linewidth]{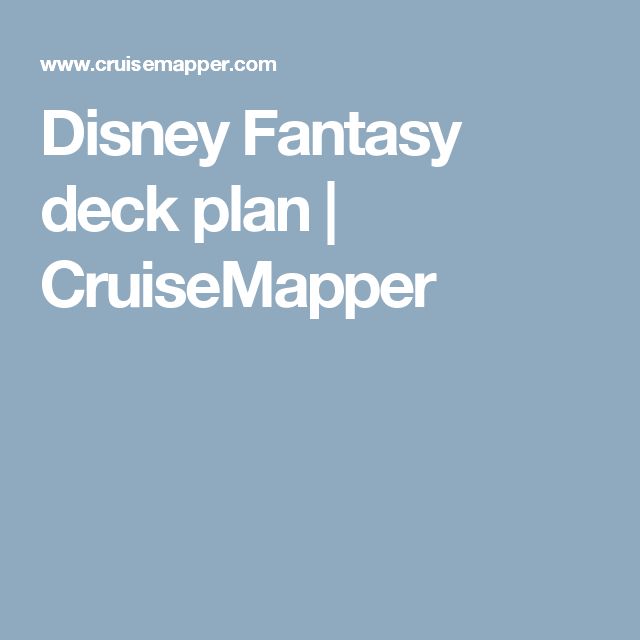}
  \caption{``Disney Fantasy deck plan | CruiseMapper''. C-f: 0.079. C: 0.401.}
  \vspace{2.2em}
  \end{subfigure}
  \begin{subfigure}{0.32\linewidth}
  \includegraphics[width=\linewidth, height=\linewidth]{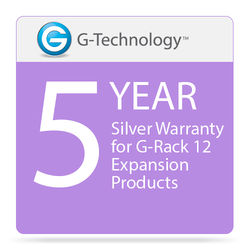}
  \caption{``G-Technology Silver 5-Year Service Warranty for G-Rack 12 Expansion Products''. C-f: 0.151. C: 0.466.}
  \end{subfigure}

  \begin{subfigure}{0.32\linewidth}
  \includegraphics[width=\linewidth, height=\linewidth]{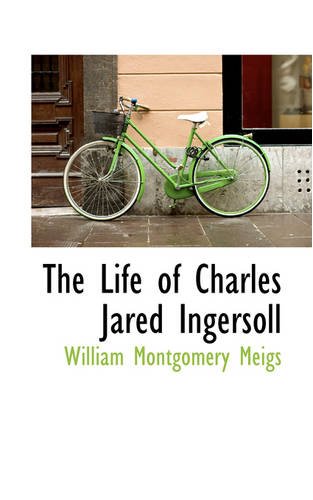}
  \caption{``The Life of Charles Jared Ingersoll: Meigs, William Montgomery''. C-f: 0.152. C: 0.404.}
  \end{subfigure}
  \begin{subfigure}{0.32\linewidth}
  \includegraphics[width=\linewidth, height=\linewidth]{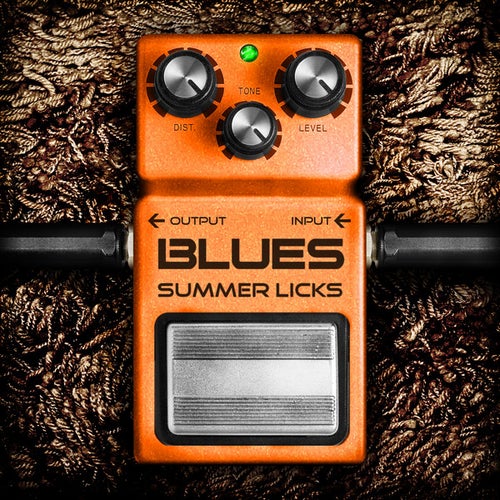}
  \caption{``BLUES: Summer Licks by Various Artists''. C-f: 0.152. C: 0.402.}
  \vspace{1.1em}
  \end{subfigure}
  \begin{subfigure}{0.32\linewidth}
  \includegraphics[width=\linewidth, height=\linewidth]{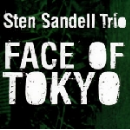}
  \caption{``sten sandell trio - face of tokyo''. C-f: 0.173. C: 0.415.}
  \vspace{2.15em}
  \end{subfigure}
  
  \caption{Samples with low flipped-CLIP scores but high CLIP scores. C-f: flipped-CLIP score. C: CLIP score.}
  \label{fig:clip_flip}
\end{figure}

In Fig.~\ref{fig:clip_flip}, we show visual examples with low flipped-CLIP scores but extremely high CLIP scores. In this case, they will be filtered out by flipped-CLIP score but not by the latter.

\subsection{CLIP v.s. flipped-CLIP}
\label{app:clip_vs_clip_flip}
\begin{figure}
    \centering
    \includegraphics[width=0.9\linewidth]{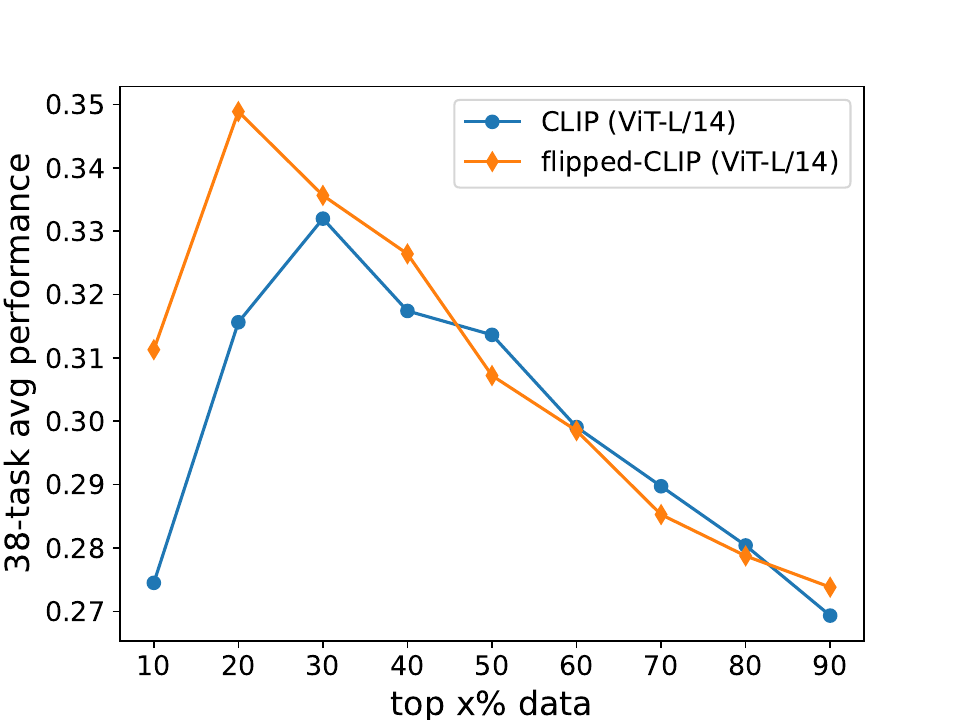}
    \caption{CLIP v.s. flipped-CLIP over 38 downstream evaluation tasks. Because each data point is based on one trained model, some variance is expected.}
    \label{fig:clip_vs_clip_flip}
\end{figure}

In Fig.~\ref{fig:clip_vs_clip_flip}, we compare performance between CLIP filtering and flipped-CLIP filtering via data bucketization, i.e., training on top $x\%$ data, where $x=10, 20, ..., 90$. We find flipped-CLIP clearly achieves superior performance when $x \leq 40$, especially in the cases of $x=10$ and $20$.

\subsection{Visual Examples with Low BLIP-ITM Scores}
\begin{figure}[t]
  \begin{subfigure}{0.32\linewidth}
  \includegraphics[width=\linewidth, height=\linewidth]{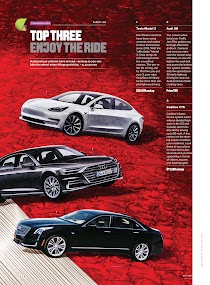}
  \caption{``WIRED- screenshot thumbnail''. B: 0.003. C-f: 0.281.}
  \end{subfigure}
  \begin{subfigure}{0.32\linewidth}
  \includegraphics[width=\linewidth, height=\linewidth]{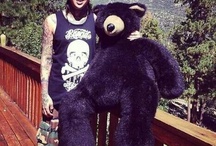}
  \caption{``Tony Perry''. B: 0.0005. C-f: 0.277.}
  \vspace{1.1em}
  \end{subfigure}
  \begin{subfigure}{0.32\linewidth}
  \includegraphics[width=\linewidth, height=\linewidth]{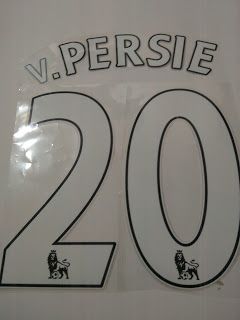}
  \caption{``NAMESET''. B: 0.002. C-f: 0.276.}
  \vspace{1.1em}
  \end{subfigure}

  \begin{subfigure}{0.32\linewidth}
  \includegraphics[width=\linewidth, height=\linewidth]{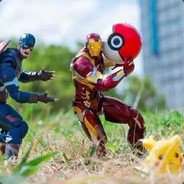}
  \caption{``Steam user avatar image''. B: 0.001. C-f: 0.274.}
  \vspace{1.1em}
  \end{subfigure}
  \begin{subfigure}{0.32\linewidth}
  \includegraphics[width=\linewidth, height=\linewidth]{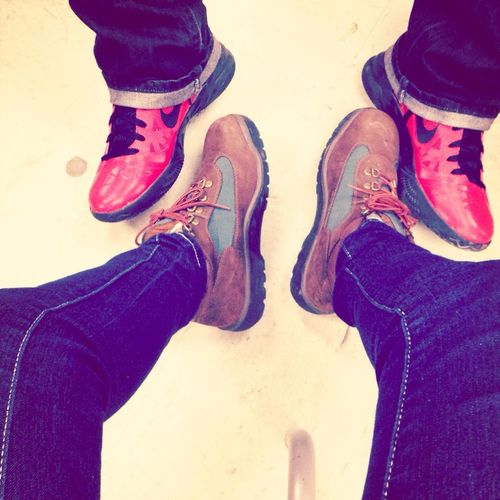}
  \caption{``kotd ! Mines Are The Beef amp; Brocs ♥✌''. B: 0.0007. C-f: 0.273.}
  \end{subfigure}
  \begin{subfigure}{0.32\linewidth}
  \includegraphics[width=\linewidth, height=\linewidth]{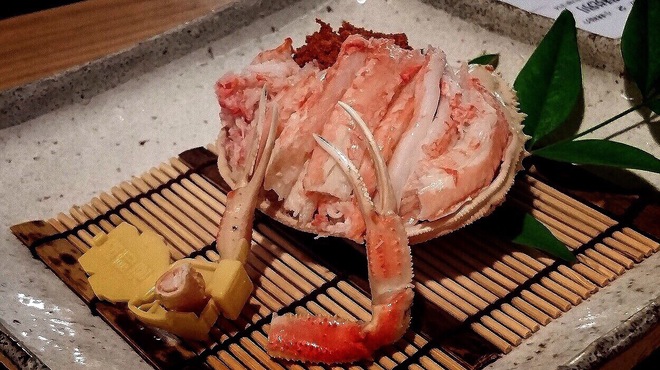}
  \caption{``越前若狭 鯖街道 - メイン写真:''. B: 0.004. C-f: 0.273.}
  \vspace{1.1em}
  \end{subfigure}
  
  \caption{Samples removed by BLIP-ITM filtering. B: BLIP-ITM score. C-f: flipped-CLIP score.}
  \label{fig:blip_itm}
\end{figure}

Fig.~\ref{fig:blip_itm} shows visual examples removed by BLIP-ITM filtering but ignored by flipped-CLIP filtering.
\label{app:removed_examples_blip_itm}

\begin{CJK}{UTF8}{ipxm}

\end{CJK}

\subsection{Implementation Details for Cluster-importance-based Data Selection}
\label{app:data_select}
\begin{table*}[t]
\centering
\begin{tabular}{c} 
\toprule
\begin{lstlisting}[language=Python]
def get_importance(center_embeddings, image_features, threshold):
    # get center to downstream similarity
    similarity = torch.einsum('ik,jk->ij', center_embeddings, image_features)
    similarity = similarity.cpu().numpy()
    
    # similarity[i,j] > threshold means image j is likely to be contained in cluster i
    valid = similarity >= threshold 
    valid = valid.astype('float')
    
    # each image is shared by how many clusters
    sums = valid.sum(axis=0, keepdims=True)
    
    # normalize the vote for each image
    importance = np.divide(valid, sums, out=np.zeros_like(valid), where=sums!=0) 
    importance = importance.sum(axis=1)

    # normalize for this task
    return importance / importance.sum()

def cluster_importance_estimation(center_embeddings, tasks, threshold):
    cluster_importance = np.zeros(center_embeddings.shape[0])
    for task in tasks:
        image_features = get_image_features(task)
        
        # get cluster importance for this task
        importance = get_importance(center_embeddings, image_features, threshold)
        
        cluster_importance += importance

    return cluster_importance / cluster_importance.sum()
\end{lstlisting}\\
\bottomrule   
\end{tabular}
\caption{Python-style pseudo code for cluster-importance estimation}
\label{alg:cluster-importance}
\end{table*}

\noindent{\textbf{Cluster importance estimation.}} We leverage $100,000$ image clusters provided by DataComp, and define the cluster importance as the proportion of downstream images that are likely to be generated/contained in each cluster. Specifically, let $\{D_j\}|_{j=1}^{n}$ be $n$ training sets of downstream tasks. 
Given a downstream training image $I_{j_k} \in D_j$, for each cluster $c_i$, we get the image-cluster similarity and compare it to a threshold: $s(i,j_k) = \mathbbm{1}(\mathrm{cos\_sim}(e_i,e_{j_k}) > thre_{imp})$, where $e_i$ and $e_{j_k}$ are the embeddings of the cluster centroid and $I_{j_k}$, respectively. Then, the importance weight of cluster $c_i$, estimated by $I_{j_k}$, is defined as $i_w(i,j_k)=s(i,j_k) / \sum_{i'}s(i',j_k)$.

Considering all images from $D_j$, the importance weight of cluster $c_i$,  estimated by $D_j$, is defined as $i_w(i,j)=\sum_{k'}i_w(i,j_{k'}) / \sum_{i',k'}i_w(i',j_{k'})$. We normalize the importance weight across all clusters as downstream training sets have unbalanced data numbers.

Finally, considering all downstream tasks that have training sets, the importance weight of cluster $c_i$, estimated by $\{D_j\}|_{j=1}^{n}$, is defined as $i_w(i)=\sum_{j'}i_w(i,j') / \sum_{i,'j'}i_w(i',j')$. We show the pseudo code in Tab.~\ref{alg:cluster-importance}.

\noindent{\textbf{Cluster-importance-based data selection (CIDS).}} Let $N$ denote the total number of data we want to sample. For cluster $c_i$, we select $N\cdot i_w(i)$ data by flipped-CLIP score. In practice, some clusters do not have enough data to meet the required number, so we collect un-selected data from all clusters, and choose the samples with top flipped-CLIP scores, to build a size-$N$ subset. We empirically choose $N$ to be $20\%$ data.

\subsection{Implementation Details for Quality-based Duplication}
\label{app:quality_dup}

In the official DataComp codebase\footnote{\url{https://github.com/mlfoundations/datacomp}}, duplicate samples are sharded next to each other, because the subset UIDs are sorted before data packing. This increases the probability that a model sees duplicate samples in a single training batch, which is harmful in contrastive learning. To avoid this issue, we intentionally pack the duplicate samples into different .npy UID files, and thus different subsets. During training, we feed concatenated subset paths to the model trainer.

\subsection{Cluster Imbalance after Importance Selection}
\label{app:semantic_dedup_cluster_imbalance}
\begin{figure}[t]
    \centering
    \includegraphics[width=0.8\linewidth]{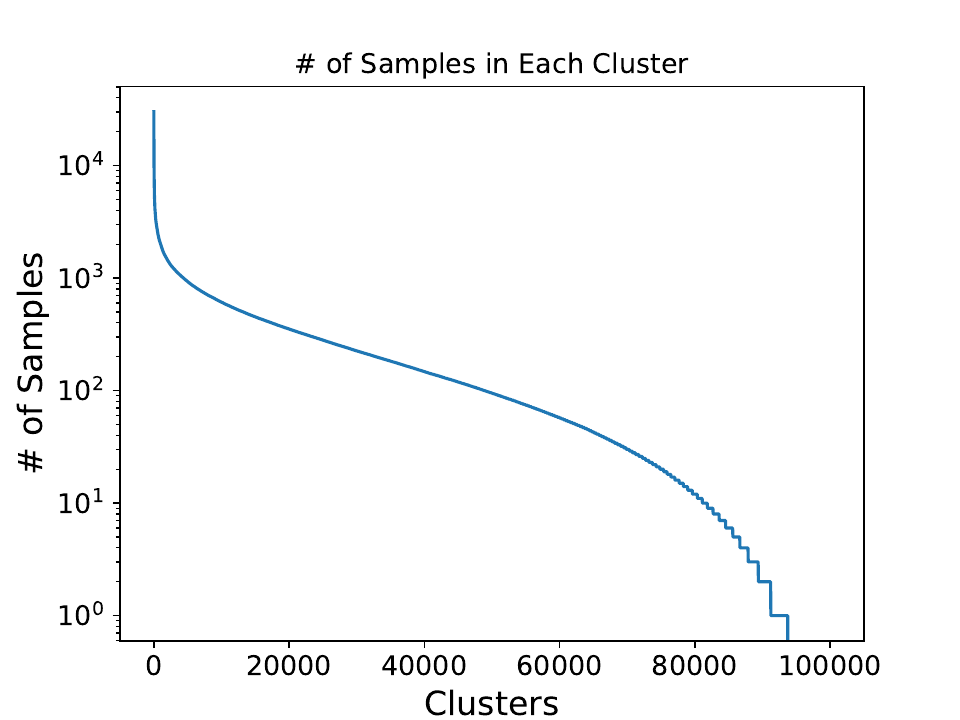}
    \caption{Numbers of samples over different clusters from cluster-importance-based data selection.}
    \label{fig:cluster_sample_dist}
\end{figure}

We visualize the number of samples in each cluster obtained in the CIDS step in Fig.~\ref{fig:cluster_sample_dist}.

\subsection{More Details on Semantic Deduplication}
\label{app:semdedup}
\begin{figure}[t]
  \begin{subfigure}{0.32\linewidth}
  \includegraphics[width=\linewidth, height=\linewidth]{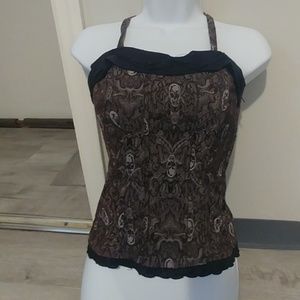}
  \end{subfigure}
 \begin{subfigure}{0.32\linewidth}
  \includegraphics[width=\linewidth, height=\linewidth]{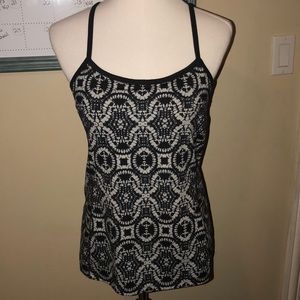}
  \end{subfigure}
   \begin{subfigure}{0.32\linewidth}
  \includegraphics[width=\linewidth, height=\linewidth]{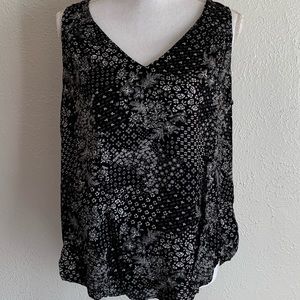}
  \end{subfigure}

  \begin{subfigure}{0.32\linewidth}
  \includegraphics[width=\linewidth, height=\linewidth]{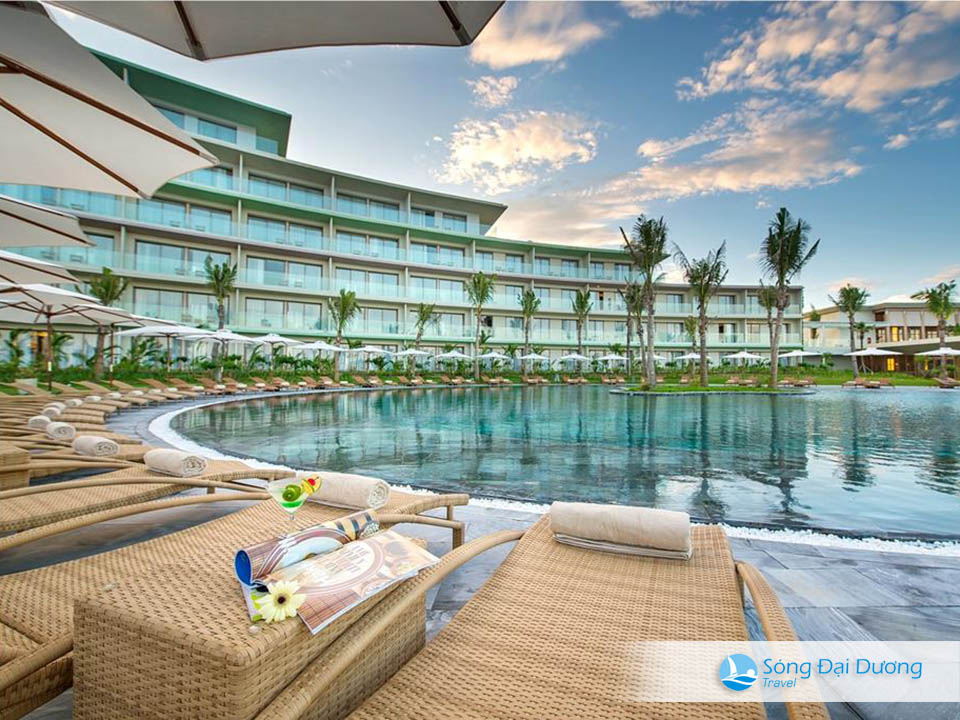}
  \end{subfigure}
 \begin{subfigure}{0.32\linewidth}
  \includegraphics[width=\linewidth, height=\linewidth]{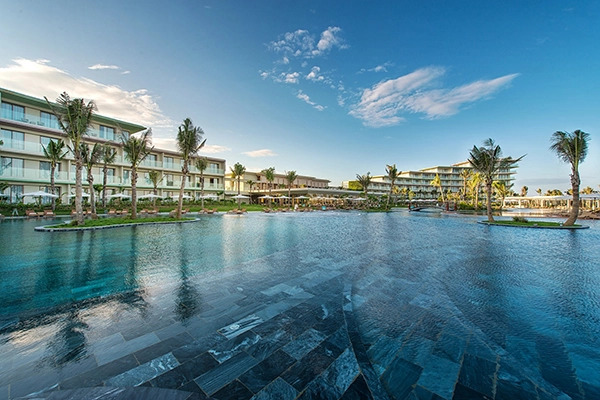}
  \end{subfigure}
  \begin{subfigure}{0.32\linewidth}
  \includegraphics[width=\linewidth, height=\linewidth]{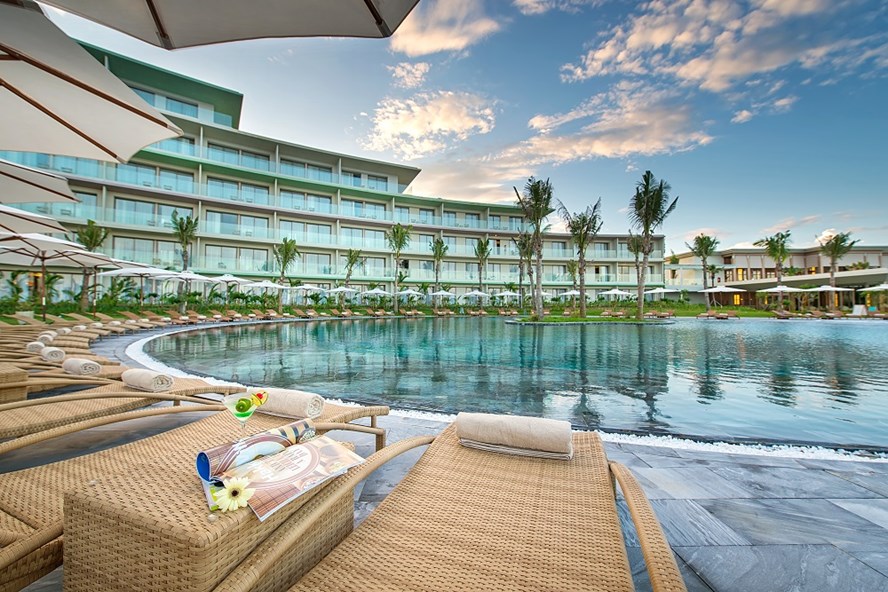}
  \end{subfigure}

  \begin{subfigure}{0.32\linewidth}
  \includegraphics[width=\linewidth, height=\linewidth]{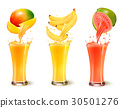}
  \end{subfigure}
  \begin{subfigure}{0.32\linewidth}
  \includegraphics[width=\linewidth, height=\linewidth]{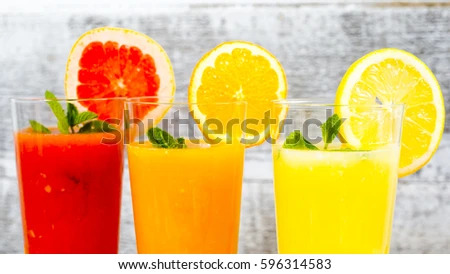}
  \end{subfigure}
  \begin{subfigure}{0.32\linewidth}
  \includegraphics[width=\linewidth, height=\linewidth]{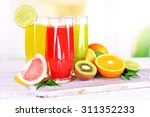}
  \end{subfigure}
  
  \caption{Visual examples of semantic deduplication. Samples in a row come from the same group, and only one image will be retained in each group. C-f: flipped-CLIP score. N: number of samples within the group.}
  \label{fig:semdedup}
\end{figure}

In Fig.~\ref{fig:semdedup}, we present visual examples from image groups of semantic deduplication, where each row is from the same group.

In addition to only using image embeddings for k-means clustering, we also tried image-text joint embeddings. Specially, we concatenate image embedding $e_i$ and text embedding $e_t$ scaled by $w_t$ to obtain joint embedding $e_j = [e_i; w_t * e_t]$. We experiment with various $w_t$ and present the best choice in Tab.~\ref{tab:semdedup_img_vs_joint}. All the results are obtained by applying semantic deduplication after \emph{single-modality filtering}, \emph{cross-modality filtering}, and image-based filtering from \cite{datacomp}. For image-based filtering, we use all available 22 training sets from downstream tasks instead of ImageNet only as in \cite{datacomp}. The experimental results indicate joint embedding does not bring improvement over image embedding only. This may be attributed to the inherent noisiness present in the DataComp captions.

\begin{table}[t]
    \centering
    \small
    \begin{tabular}{l|c|c|c}
    \bottomrule
        Emb. type  & Dataset size & 38 tasks & IN\\
    \midrule
        Image emb. & 22.4M & 0.3527 & 0.3208\\
        Joint emb. ($w_t = 0.25$) & 22.7M & 0.3504 & 0.3222\\
    \bottomrule
    \end{tabular}
    \caption{Comparison between different embeddings in k-means clustering of semantic deduplication.}
    \label{tab:semdedup_img_vs_joint}
\end{table}

\subsection{Digit Recognition Enhancement}
\label{app:added_examples_blip2}

For digit recognition enhancement, we create two textual prompts:

\begin{itemize}
    \setlength\itemsep{0em}
    \item ``Question: Does this image contain a number or a digit, Answer in yes or no only? Answer:"
    \item ``Question: Does this image contain a digit between 0-10, Answer in yes or no only? Answer:"
\end{itemize}

\begin{figure}[t]
  \begin{subfigure}{0.24\linewidth}
  \includegraphics[width=\linewidth, height=\linewidth]{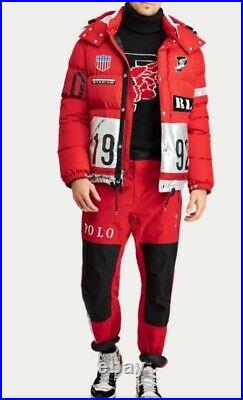}
  \caption{$1, 9$}
  \end{subfigure}
  \begin{subfigure}{0.24\linewidth}
  \includegraphics[width=\linewidth, height=\linewidth]{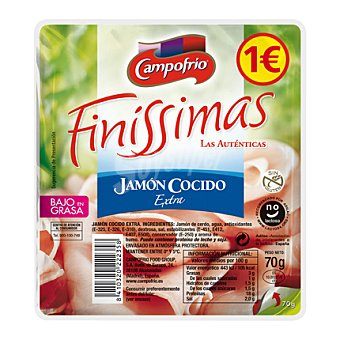}
  \caption{$1$}
  \end{subfigure}
  \begin{subfigure}{0.24\linewidth}
   \includegraphics[width=\linewidth, height=\linewidth]{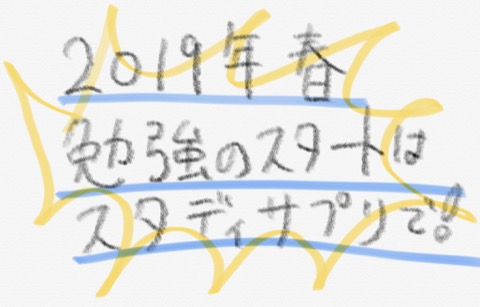}
  \caption{$2, 0, 1, 9$}
  \end{subfigure}
  \begin{subfigure}{0.24\linewidth}
  \includegraphics[width=\linewidth, height=\linewidth]{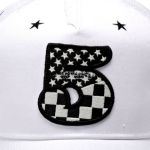}
  \caption{$5$.}
  \end{subfigure}
  \begin{subfigure}{0.24\linewidth}
   \includegraphics[width=\linewidth, height=\linewidth]{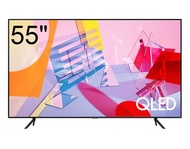}
  \caption{$5, 5$}
  \end{subfigure}
  \begin{subfigure}{0.24\linewidth}
  \includegraphics[width=\linewidth, height=\linewidth]{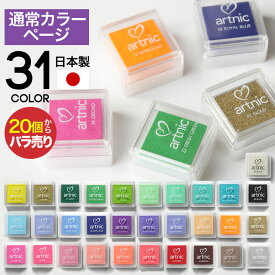}
  \caption{$3, 1$}
  \end{subfigure}
  \begin{subfigure}{0.24\linewidth}
  \includegraphics[width=\linewidth, height=\linewidth]{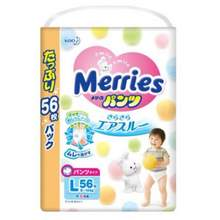}
  \caption{$5, 6$}
  \end{subfigure}
  \begin{subfigure}{0.24\linewidth}
  \includegraphics[width=\linewidth, height=\linewidth]{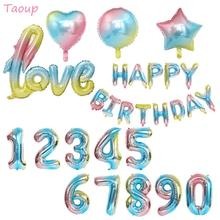}
  \caption{$0$-$9$}
  \end{subfigure}
  \caption{Images detected as useful for digit classification. The observed digits are listed in the sub-captions.}
  \label{fig:BLIP2_qualitative}
\end{figure}

We provide these prompts along with images to the BLIP2 model and generate a textual response. If the BLIP2 model outputs ``Yes`` to both questions for a given image then that image is classified as useful for digit recognition. Fig.~\ref{fig:BLIP2_qualitative} provides examples of such images.

\begin{table}[t]
    \centering
    \footnotesize
    \begin{tabular}{l|c|c|c}
    \bottomrule
        Model type  & Dataset size & 38 tasks & IN\\
    \midrule
        CLIP ViT-L/14~\cite{radford2021learning} & 24.1M & 0.3495 & 0.3173 \\
        EVA02-CLIP-bigE-14-plus~\cite{EVA-CLIP} & 24.1M & 0.3353 & 0.2954 \\
    \bottomrule
    \end{tabular}
    \caption{Comparison between CLIP models for \emph{cross-modality filtering}.}
    \label{tab:clip_vs_evaclipe}
\end{table}

\subsection{Non-overlapping Examples between the Best Official DataComp Subset and Ours}
\label{app:ours_vs_datacomp}

\begin{figure*}[t]
\centering
\begin{subfigure}[t]{0.15\textwidth}
    \makebox[0pt][r]{\makebox[30pt]{\raisebox{40pt}{\rotatebox[origin=c]{90}{\small Exclusive to DC}}}}%
    \includegraphics[width=\textwidth, height=\textwidth]
    {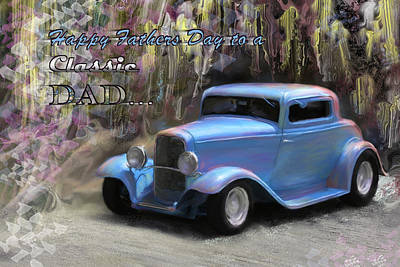}
    \caption{``Fathers Day Classic Dad Poster by Susan Kinney''}
    \makebox[0pt][r]{\makebox[30pt]{\raisebox{40pt}{\rotatebox[origin=c]{90}{\small Exclusive to DC}}}}%
    \includegraphics[width=\textwidth, height=\textwidth]
    {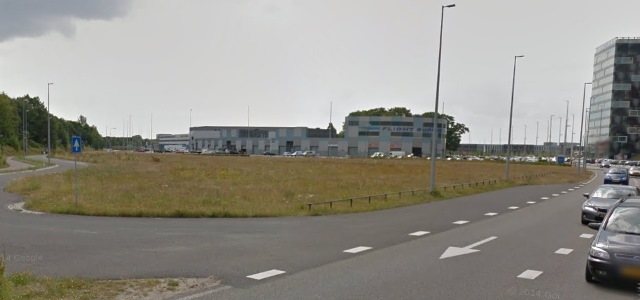}
    \caption{``Perceel Tankstation Eindhoven Airport''}
    \vspace{2.2em}
    \makebox[0pt][r]{\makebox[30pt]{\raisebox{40pt}{\rotatebox[origin=c]{90}{\small Exclusive to Ours}}}}%
    \includegraphics[width=\textwidth, height=\textwidth]
    {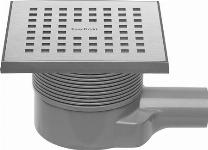}
    \caption{``Easy Drain Aqua 150x150mm afvoerput RVS geborsteld MSI-6''}
    \vspace{2.2em}
    \makebox[0pt][r]{\makebox[30pt]{\raisebox{40pt}{\rotatebox[origin=c]{90}{\small Exclusive to Ours}}}}%
    \includegraphics[width=\textwidth, height=\textwidth]
    {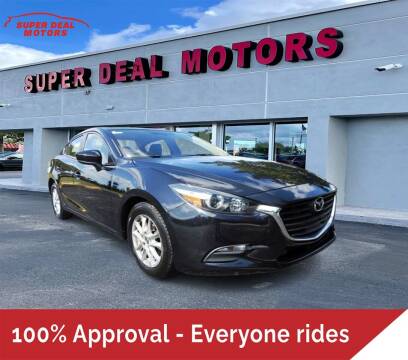}
    \caption{``2017 Mazda MAZDA3 for sale at SUPER DEAL MOTORS in Hollywood FL''}
\end{subfigure}
\begin{subfigure}[t]{0.15\textwidth}
    \includegraphics[width=\textwidth, height=\textwidth, height=\textwidth]  
    {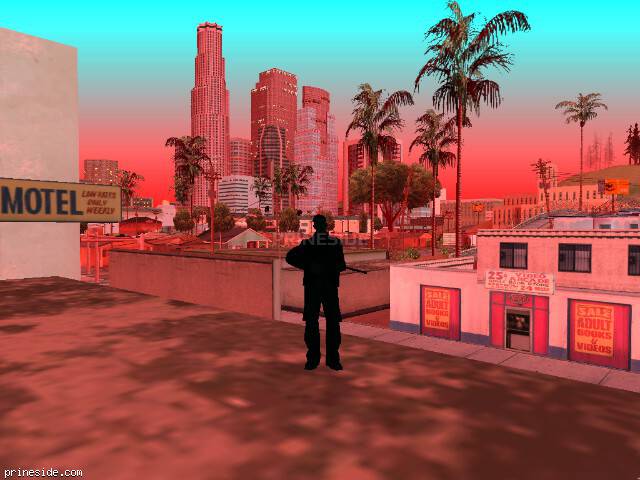}
    \caption{``GTA San Andreas weather ID 484 at 20 hours''}
    \includegraphics[width=\textwidth, height=\textwidth]
    {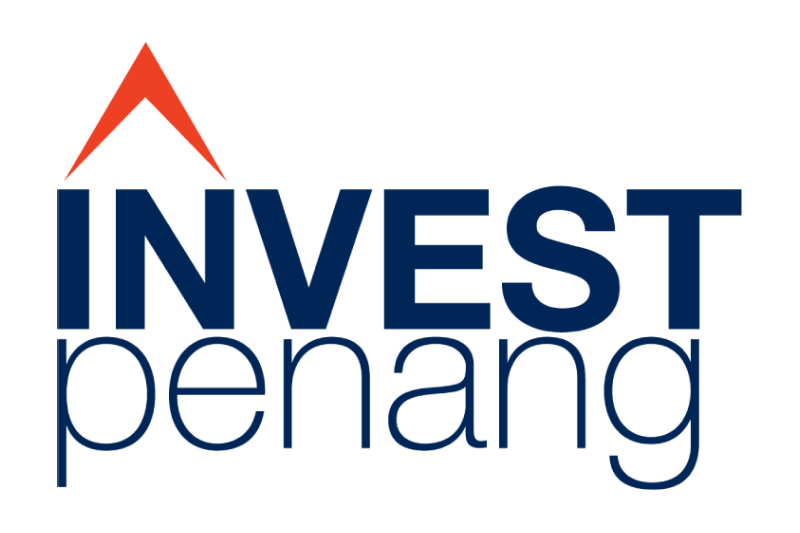}
    \caption{``Penang Dialogue: Penang ManPower''}
    \vspace{2.2em}
    \includegraphics[width=\textwidth, height=\textwidth]  
    {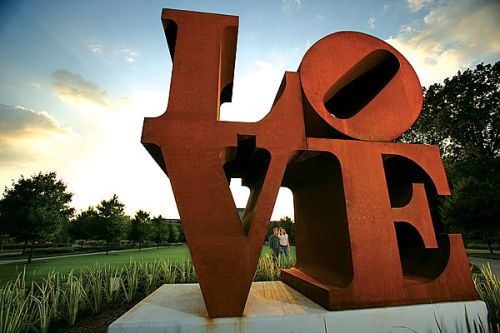}
    \caption{``LOVE sculpture, Robert Indiana, 1970 -\&nbsp; Indianapolis Museum of Art''}
    \vspace{1.1em}
    \includegraphics[width=\textwidth, height=\textwidth]
    {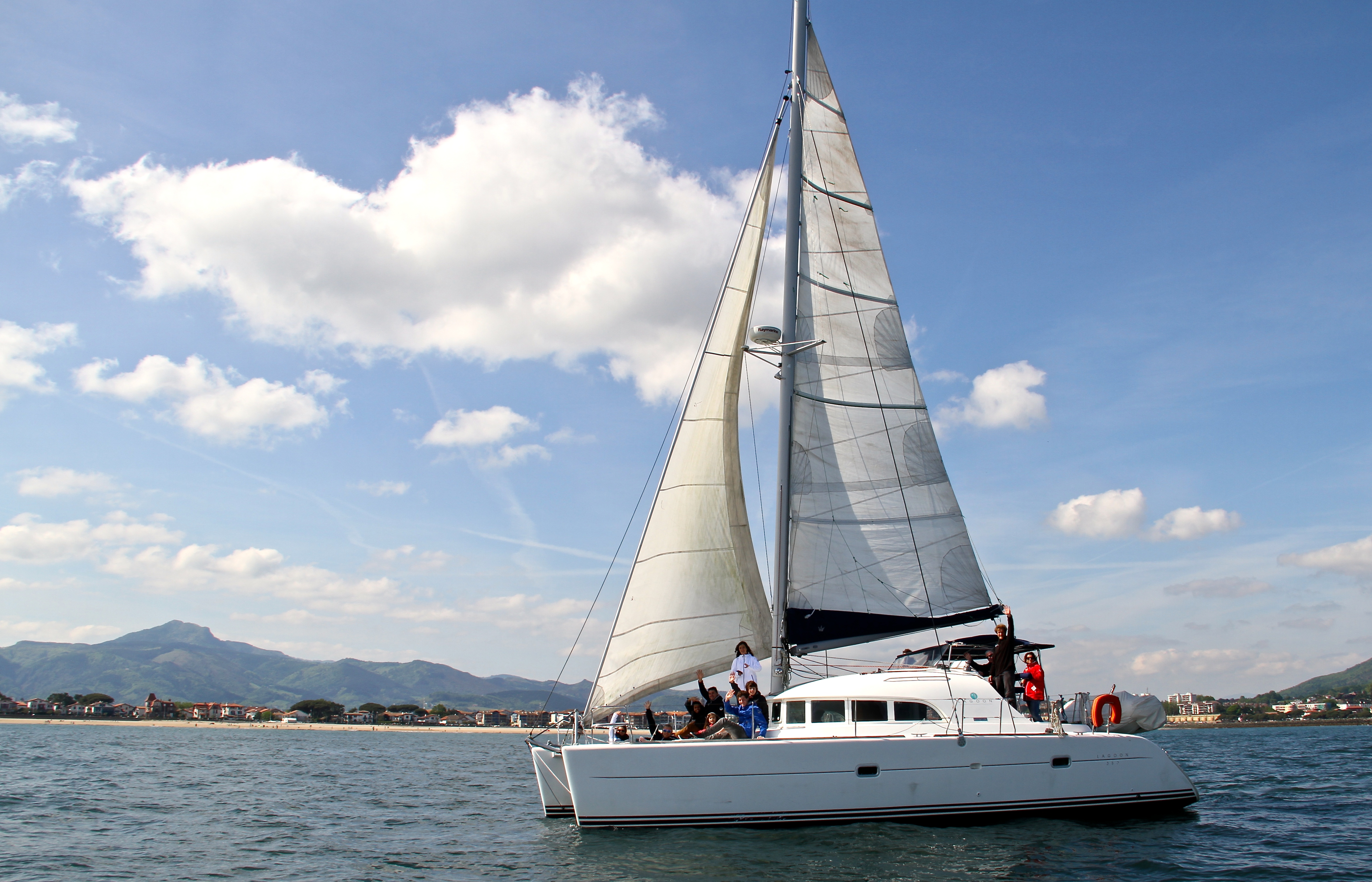}
    \caption{``Dégustation de vins sur un catamaran dans la baie de Saint Jean de Luz''}
\end{subfigure}
\begin{subfigure}[t]{0.15\textwidth}
    \includegraphics[width=\textwidth, height=\textwidth]  
    {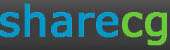}
    \caption{``ShareCG home''}
    \vspace{2.2em}
    \includegraphics[width=\textwidth, height=\textwidth]
    {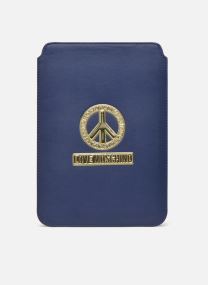}
    \caption{``Wallets \&amp; cases Bags Ipad clutch''}
    \vspace{2.2em}
    \includegraphics[width=\textwidth, height=\textwidth]  
    {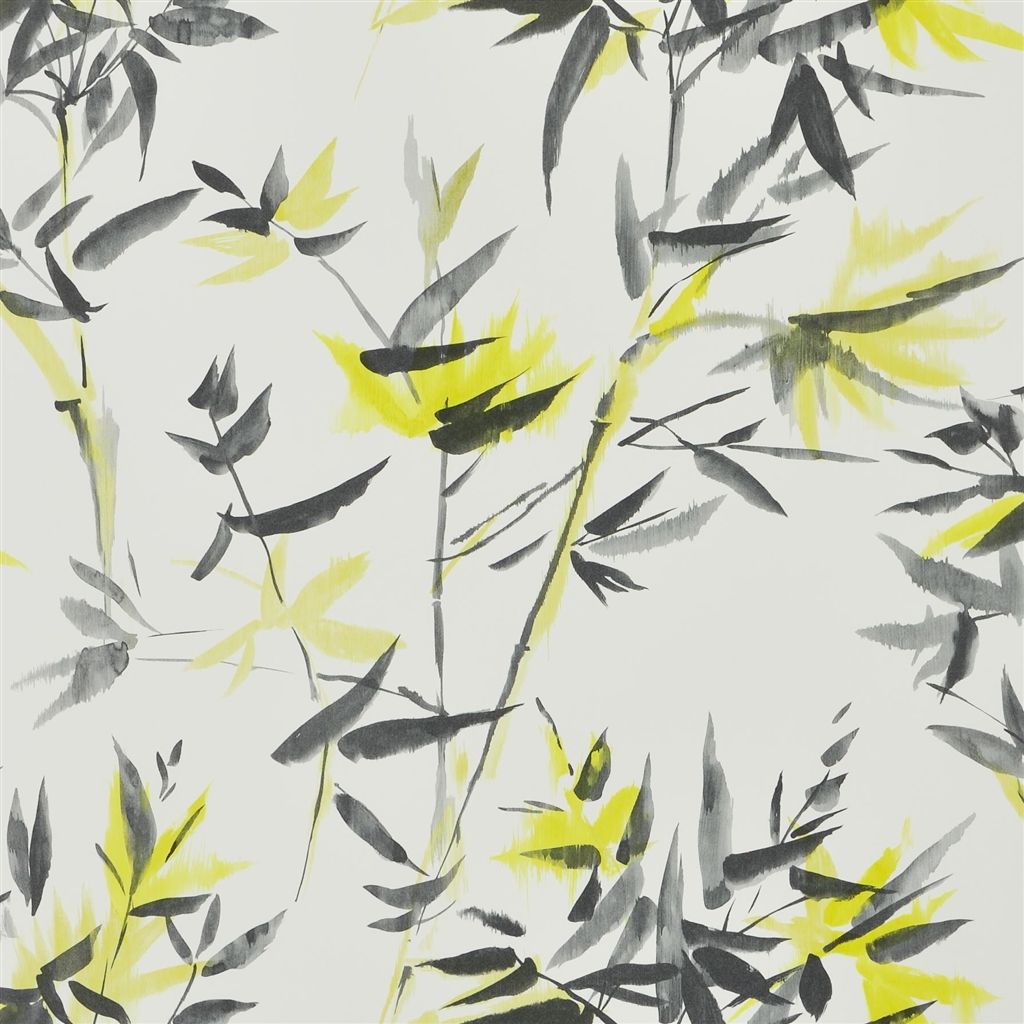}
    \caption{``Wallpaper - Designers Guild - Shanghai Garden - Bamboo-Acacia - Half drop - 68.5 cm x 10 m''}
    \includegraphics[width=\textwidth, height=\textwidth]
    {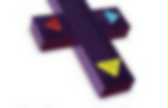}
    \caption{``background preloader''}
\end{subfigure}
\begin{subfigure}[t]{0.15\textwidth}
    \includegraphics[width=\textwidth, height=\textwidth]  
    {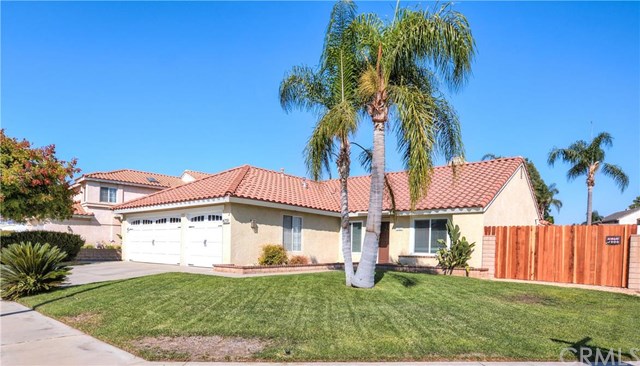}
    \caption{``7441 Plumaria Drive, Fontana, CA 92336''}
    \includegraphics[width=\textwidth, height=\textwidth]
    {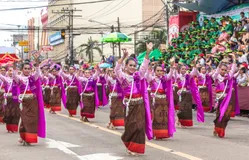}
    \caption{``Thai dance at traditional candle procession festival of Buddha Royalty Free Stock Photos''}
    \includegraphics[width=\textwidth, height=\textwidth]  
    {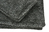}
    \caption{``Плед серо-черный Fur Blanket Graphite''}
    \vspace{2.2em}
    \includegraphics[width=\textwidth, height=\textwidth]
    {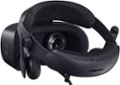}
    \caption{``Alt View Zoom 14. Samsung - HMD Odyssey Virtual Reality Headset for Compatible Windows PCs.''}
\end{subfigure}
\begin{subfigure}[t]{0.15\textwidth}
    \includegraphics[width=\textwidth, height=\textwidth]  
    {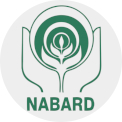}
    \caption{``NABARD Grade B 2021 Mock Test 3 Phase I''}
    \includegraphics[width=\textwidth, height=\textwidth]
    {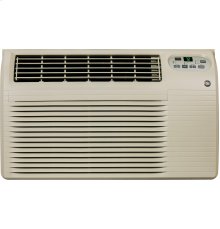}
    \caption{``GE\&reg; 230/208 Volt Built-In Heat/Cool Room Air Conditioner''}
    \vspace{1.1em}
    \includegraphics[width=\textwidth, height=\textwidth]  
    {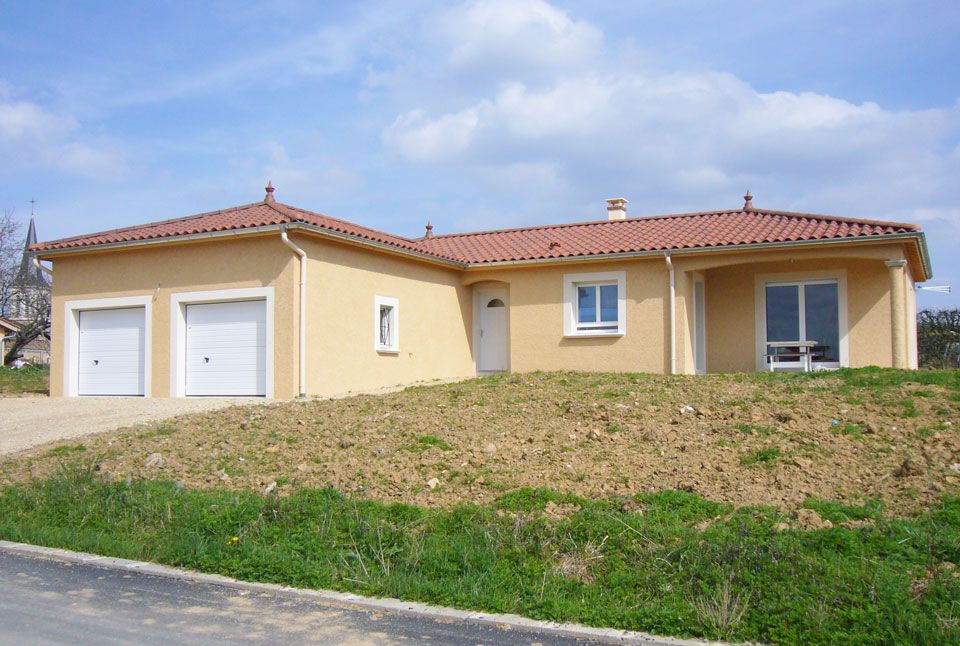}
    \caption{``maison plain pied rhone alpes''}
    \vspace{4.4em}
    \includegraphics[width=\textwidth, height=\textwidth]
    {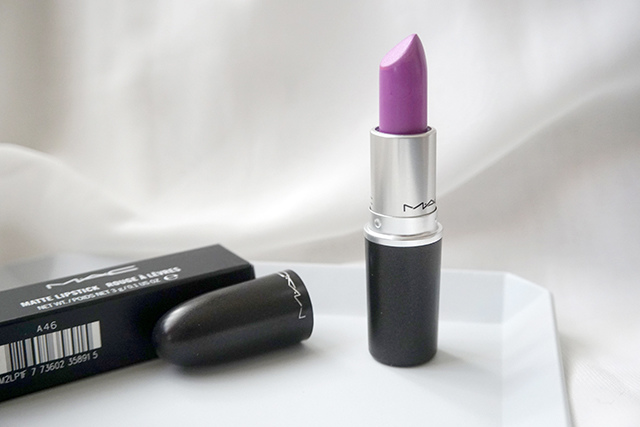}
    \caption{``MAC lipstick霧幻性感唇膏Lavender Jade \&amp;閃亮星澤唇膏Angel 16.JPG''}
\end{subfigure}
\begin{subfigure}[t]{0.15\textwidth}
    \includegraphics[width=\textwidth, height=\textwidth]  
    {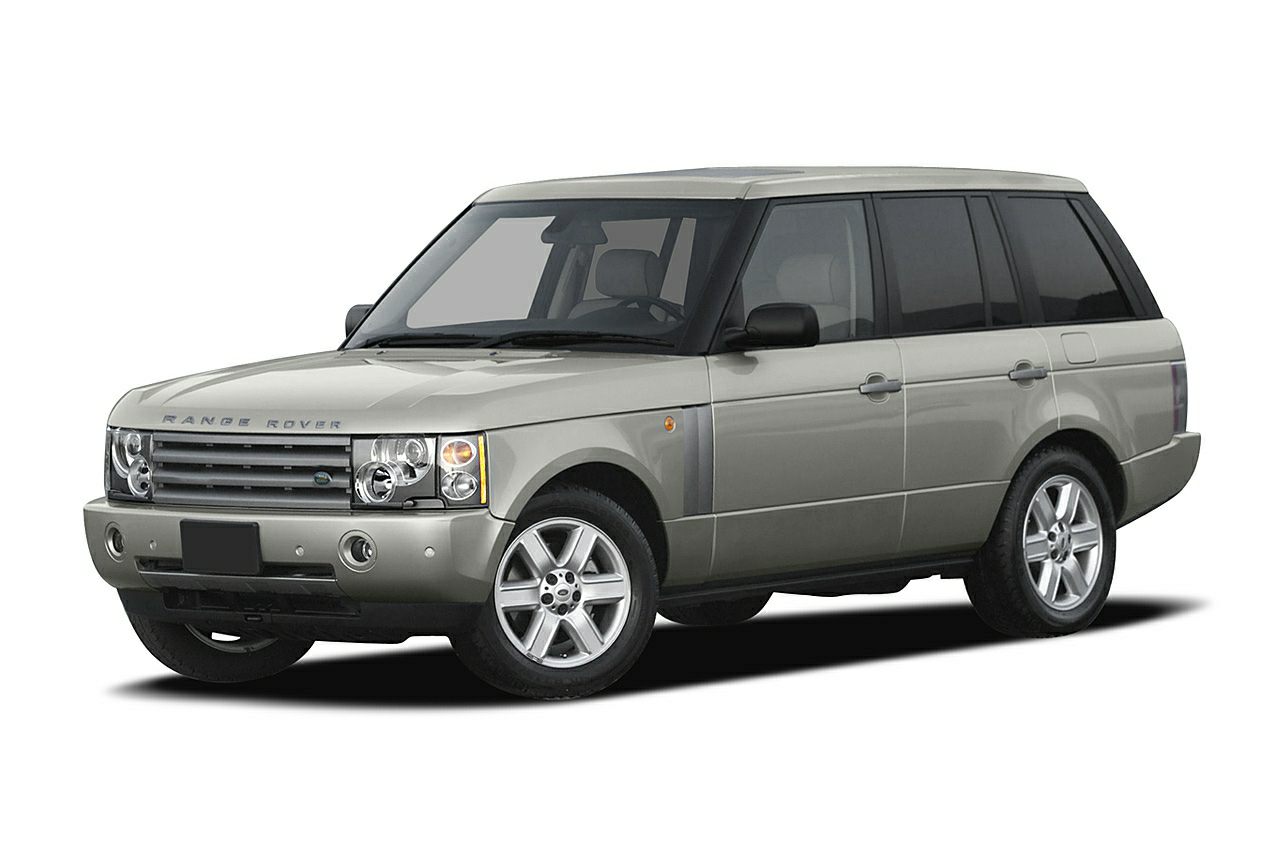}
    \caption{``2004 Land Rover Range Rover''}
    \vspace{1.1em}
    \includegraphics[width=\textwidth, height=\textwidth]
    {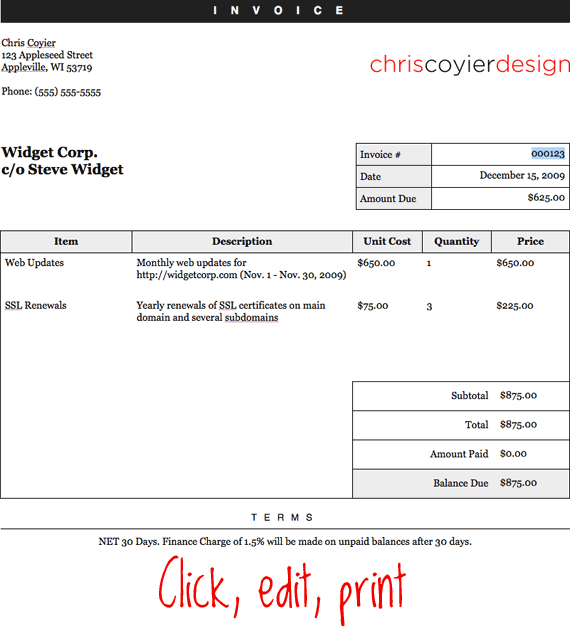}
    \caption{``Pigbrotherus Terrific Make An Editableprintable Html Invoice'' (truncated)}
    \includegraphics[width=\textwidth, height=\textwidth]  
    {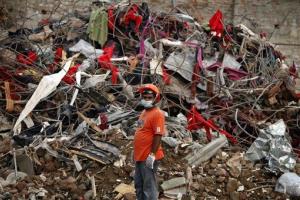}
    \caption{``A Bangladeshi rescuer stands amid the rubble during search and recovery efforts Sunday.''}
    \vspace{1.1em}
    \includegraphics[width=\textwidth, height=\textwidth]
    {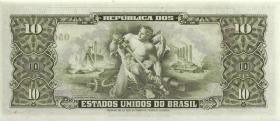}
    \caption{``Brasilien / Brazil P.177a 10 Cruzeiros (1962) (1)''}
\end{subfigure}
\caption{Visual examples exclusive to ``Image-based $\cap$ CLIP score (L/14 30\%)''~\cite{datacomp} and to our best-performed subset.}
\label{fig:ours_vs_datacomp}
\end{figure*}

In Fig.~\ref{fig:ours_vs_datacomp}, we present a qualitative comparison between the data samples selected by the best-performing official DataComp subset and our proposed method. Specifically, we display randomly sampled visual examples that are exclusive to ``Image-based $\cap$ CLIP score (L/14 30\%)" or our top-performing resulting subset (last row in Tab.~\ref{tab:results}). While visually assessing and comparing the distributions of these two subsets can be challenging, it is evident that our subset comprises fewer textual images, which can be attributed to the use of flipped-CLIP score.

\begin{CJK}{UTF8}{ipxm}

\end{CJK}

\subsection{CLIP Model Selection}
\label{app:clipb_clipl_evaclipe}
In our flipped-CLIP filtering approach, we also conducted experiments with the EVA02-CLIP-bigE-14-plus model~\cite{EVA-CLIP}. It is worth noting that this model was trained without the use of horizontal flipping as a data augmentation technique. We compare it with CLIP ViT-L/14~\cite{radford2021learning} after the application of \emph{single-modality filtering}, \emph{cross-modality filtering}, and image-based filtering~\cite{datacomp}, following the same configuration as described in App.~\ref{app:semdedup}. The results, as presented in Tab.~\ref{tab:clip_vs_evaclipe}, reveal that EVA02-CLIP-bigE-14-plus does not surpass the performance of CLIP ViT-L/14, despite the former demonstrating superior prediction accuracy. This discrepancy leads us to hypothesize that a model's performance may not be correlated with its ability for data filtering. Investigating the underlying reasons for this observation remains a potential avenue for future research.

\end{document}